\documentclass{article}

    \usepackage[preprint]{neurips_2025}

\usepackage[utf8]{inputenc} % allow utf-8 input
\usepackage[T1]{fontenc}    % use 8-bit T1 fonts
\usepackage{hyperref}       % hyperlinks
\usepackage{url}            % simple URL typesetting
\usepackage{booktabs}       % professional-quality tables
\usepackage{amsfonts}       % blackboard math symbols
\usepackage{nicefrac}       % compact symbols for 1/2, etc.
\usepackage{microtype}      % microtypography
\usepackage{xcolor}         % colors
\usepackage{graphicx}
\usepackage{amsmath}
\usepackage{comment}
\usepackage{tabularx}
\usepackage{makecell}
\usepackage{multirow}
\usepackage{wrapfig,lipsum,booktabs}

\usepackage{xspace}

\newcommand{\etal}{\textit{et al.}}

\title{$\varphi$-Adapt: A Physics-Informed Adaptation Learning Approach to 2D Quantum Material Discovery}

\author{%
  Hoang-Quan Nguyen$^1$, Xuan Bac Nguyen$^1$, Sankalp Pandey$^1$, \\
  \textbf{Tim Faltermeier$^2$, Nicholas Borys$^2$, Hugh Churchill$^3$, Khoa Luu$^1$} \\
  $^1$Department of Electrical Engineering and Computer Science, University of Arkansas, AR\\
  $^2$Department of Physics, Montana State University, MT\\
  $^3$Department of Physics, University of Arkansas, AR\\
  \texttt{\{hn016, xnguyen, sankalpp, hchurch, khoaluu\}@uark.edu} \\
  \texttt{\{timfaltermeier, nicholas.borys\}@montana.edu}
}

\begin{document}

\maketitle

\begin{abstract}
Characterizing quantum flakes is a critical step in quantum hardware engineering because the quality of these flakes directly influences qubit performance. Although computer vision methods for identifying two-dimensional quantum flakes have emerged, they still face significant challenges in estimating flake thickness. These challenges include limited data, poor generalization, sensitivity to domain shifts, and a lack of physical interpretability. In this paper, we introduce one of the first Physics-informed Adaptation Learning approaches to overcome these obstacles. We focus on two main issues, i.e., data scarcity and generalization. First, we propose a new synthetic data generation framework that produces diverse quantum flake samples across various materials and configurations, reducing the need for time-consuming manual collection. Second, we present $\varphi$-Adapt, a physics-informed adaptation method that bridges the performance gap between models trained on synthetic data and those deployed in real-world settings. Experimental results show that our approach achieves state-of-the-art performance on multiple benchmarks, outperforming existing methods.
Our proposed approach advances the integration of physics-based modeling and domain adaptation. It also addresses a critical gap in leveraging synthesized data for real-world 2D material analysis, offering impactful tools for deep learning and materials science communities.
\end{abstract}

\section{Introduction}

The discovery of two-dimensional (2D) material flakes is a critical step toward constructing Van der Waals heterostructures, which enable a wide range of applications and fundamental studies, including those in quantum mechanics \cite{james2021recent,lemme20222d,liu2022chemical}. However, the natural randomness in shapes and spatial distributions of the quantum flakes makes their identification and exploration challenging. This process typically relies on repetitive, manual optical microscopy searches, significantly limiting the complexity and scalability of heterostructure fabrication. Furthermore, estimating the thickness of identified flakes adds another layer of difficulty, as it often requires transferring samples to an Atomic Force Microscope (AFM) for measurement, which effectively doubles the manual effort involved. Motivated by these challenges, automating the detection of exfoliated 2D material flakes via deep learning approaches has emerged in recent years. 

\begin{figure}
\centering
\includegraphics[width=0.9\linewidth]{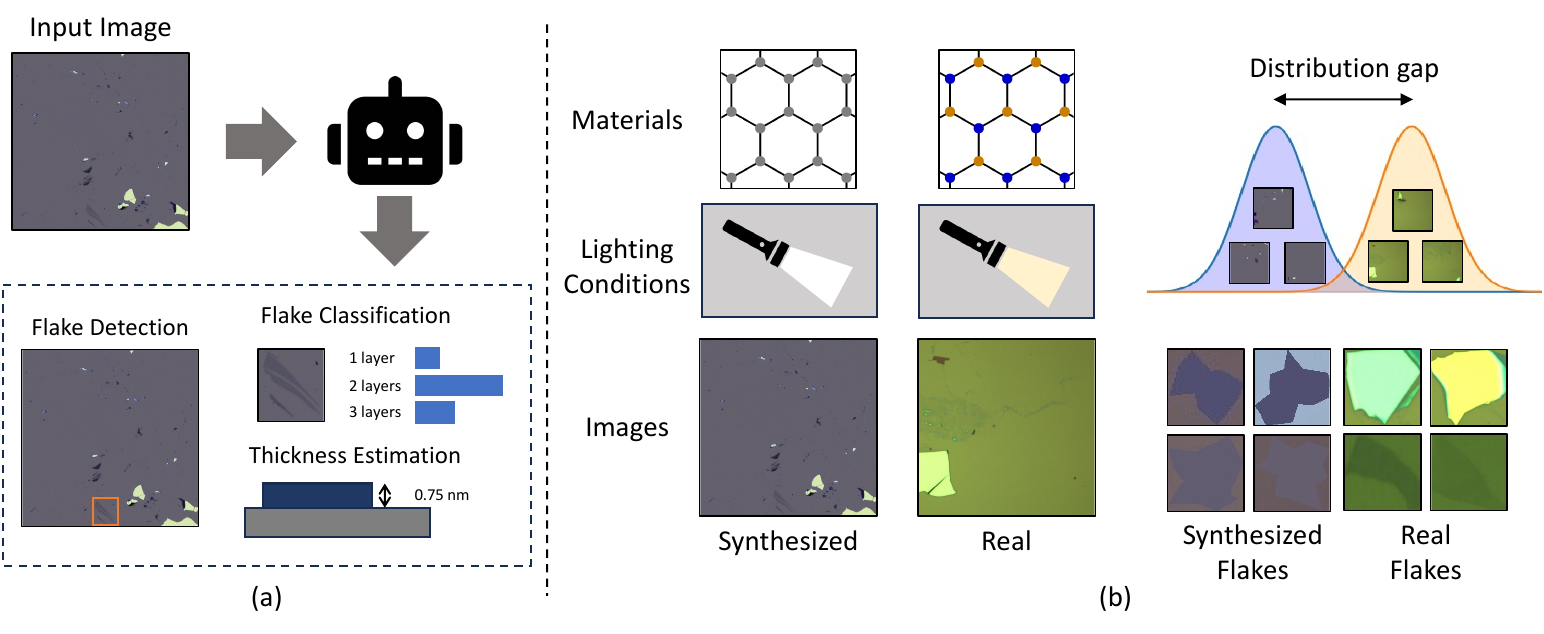}
\vspace{-4mm}
\caption{
(a) The 2D material identification tasks. 
(b) Due to the differences between experimental settings, e.g., materials and lighting, specific flakes have different appearances that make the distribution gaps between the synthesized and real datasets and challenge the 2D material identification tasks.
}
\label{fig:problem_definition}
\vspace{-6mm}
\end{figure}

\noindent
\textbf{The Challenges of 2D Flake Identification.} The critical challenge in the identification of a 2D flake is to determine the thickness of that flake. Firstly, 2D materials are often just a few atomic layers thick, sometimes even monolayers. Distinguishing monolayer, bilayer, and few-layer flakes requires sub-nanometer resolution, which is beyond the capabilities of typical optical methods. Secondly, under a regular optical microscope, flakes have very subtle contrast differences with the substrate (e.g., SiO$_2$/Si). The flakes with different thicknesses, such as monolayer and bilayer, often appear visually similar in optical images. The pixel-level contrast between different thicknesses can be subtle and non-linear. Lastly, contrast can vary with wavelength, substrate thickness, and flake material, making it non-trivial to visually identify the number of layers.

\noindent
Apart from typical depth estimation problems, flake thickness estimation focuses on determining the physical thickness of microscopic two-dimensional (2D) materials, such as graphene or MoS$_2$, often at nanometer to micrometer scales. It heavily depends on material-specific properties like optical contrast, interference patterns, and spectral signatures, requiring precise tools such as optical microscopy, Raman spectroscopy, or atomic force microscopy. In contrast, depth estimation aims to recover the geometric distance from the camera to objects in a 3D scene, operating at much larger, real-world scales. It typically relies on geometric cues, stereo matching, or depth sensors, and is mainly independent of the material properties. While flake thickness estimation demands sensitivity to subtle visual variations under high magnification, depth estimation involves modeling large-scale scene structure and dealing with occlusion, textureless surfaces, and lighting variation. Consequently, thickness estimation is highly specialized and material-dependent, whereas depth estimation is more generalized and geometry-driven. Thus, applying depth estimation approaches recently is not appropriate for the thickness estimation of flakes.

\noindent
\textbf{Limitations of Prior Work.} Prior works on flake thickness estimation from optical microscopy images face significant limitations across traditional and deep learning approaches. 
Traditional methods such as SVM \cite{hearst1998support} and K-NN \cite{fix1985discriminatory} rely heavily on hand-crafted features like color histograms or texture descriptors, which often fail to capture the subtle and non-linear differences between mono-layer and few-layer flakes. 
These methods suffer from poor generalization across imaging setups and materials, are not scalable to large datasets, and lack end-to-end optimization. More recent deep learning approaches \cite{han2020deep,masubuchi2020deep} address some of these issues by learning discriminative features directly from data, but they introduce new challenges. Deep models require large labeled datasets, which are difficult to obtain due to the need for precise ground-truth from tools like AFM or Raman spectroscopy. They also struggle with domain shifts, often overfit to specific materials or imaging conditions, and lack interpretability. Furthermore, most deep models ignore the underlying optical physics of 2D materials and provide no uncertainty estimates, limiting their reliability in real-world scientific workflows.
It is important to note that previous methods \cite{masubuchi2020deep,uslu2024open} were developed and evaluated on their datasets, which raises concerns about their applicability in real-world scenarios. In practice, exfoliation systems are designed differently across research groups, resulting in significant variations in the collected data, as illustrated in Figure \ref{fig:problem_definition}. 

\textbf{Problem Motivation.} This problem must be addressed to ensure the robustness and applicability of automated exfoliation analysis systems in real-world settings. The lack of diverse, standardized datasets and high collection costs limits model generalization across different experimental setups. 
To overcome this challenge, we propose a novel approach to enhance the automatic 2D material identification systems in real-world settings. 
First, we generate synthesized data to alleviate the data scarcity and provide diverse, well-controlled training samples that simulate a wide range of exfoliation scenarios. Second, we employ a test-time domain adaptation technique to bridge the gap between the model trained on synthesized data and real-world data, enabling the model to generalize effectively across different experimental conditions. Unlike typical domain adaptation methods, our approach leverages the optical properties of materials in the adaptation step. This approach enhances the model’s robustness and provides a scalable pathway toward real-world deployment of exfoliation detection systems.

\textbf{Contributions of this Work.}
This work proposes a novel approach to analyzing 2D materials under varying image conditions. First, we develop a novel \textit{physics-informed adaptation network} that calibrates material images captured under different optical distributions.
Second, we introduce a learning strategy to capture the underlying optical properties of 2D materials effectively. To make the model robust to the target 2D materials, we propose a novel synthesized dataset to make the model learn the optical distributions of the materials and shift between these distributions easily. Furthermore, we present a new source-free domain adaptation approach via entropy minimization that enables the model to adapt to unseen material domains without requiring additional supervised training. Extensive experiments demonstrate that our method achieves state-of-the-art (SOTA) performance on flake detection and thickness estimation benchmarks.
By bridging the gap between physics-based modeling and domain adaptation, our approach empowers the use of synthesized data for real-world 2D material identification.

\section{Related Work}

\subsection{Automatic 2D Materials Identification}

Exploring 2D quantum materials has become an increasingly prominent area of focus in research, accelerating the development of quantum technologies and applications \cite{dendukuri2019defining,nguyen2023quantum}. 
Han \etal \cite{han2020deep} proposed a deep learning method for the optical identification of 13 distinct 2D material types from optical microscopic images. 
Their described architecture employed multiple convolutions, batch normalization \cite{ioffe2015batch}, ReLU activations \cite{agarap2018deep}, and a final softmax layer. 
This approach was motivated by the UNet architecture \cite{ronneberger2015u} for segmenting flakes of varying thickness. 
Masubuchi \etal \cite{masubuchi2020deep} presented an end-to-end pipeline for 2D flake identification. 
Their work utilized Mask R-CNN \cite{he2017mask} to predict the bounding boxes and segmentation masks of individual flakes. 
Their method processes optical microscopy images as input and outputs the location and material classification of each flake. 
Nguyen \etal \cite{nguyen2024two} utilized self-attention and soft-labeling to identify the flake material.
Additionally, false negative object detection was introduced \cite{luu2024automatically}.
While the deep learning methods require a large amount of training data, the machine learning paradigm can be applied, e.g., Gaussian mixture models for flake classification \cite{uslu2024open}. 
However, machine learning methods often perform poorly when applied to noisy datasets. With the rapid growth of foundation models, zero-shot and few-shot learning can be utilized for the 2D material identification \cite{kirillov2023segment,ravi2024sam}.
However, the physical knowledge of the 2D material should be considered for better performance.

\subsection{Unsupervised Domain Adaptation}

Unsupervised Domain Adaptation (UDA) is an extensively studied research topic in transfer learning, particularly relevant for tasks where annotated data in the target domain is scarce. 
In semantic segmentation, the goal of UDA methods is to transfer knowledge from a labeled source domain to an unlabeled target domain. 
Traditional UDA methods consist of domain discrepancy minimization \cite{ganin2015unsupervised, long2015learning, tzeng2017adversarial}, adversarial learning \cite{chen2018road,chen2017no, hoffman2018cycada, hoffman2016fcns, hong2018conditional, tsai2018learning}, entropy minimization \cite{murez2018image, pan2020unsupervised, vu2019advent, zhu2017unpaired, nguyen2022self, truong2023comal}, and self-training \cite{zou2018unsupervised}.
Several UDA approaches improve performance by using the privileged information available in the source data \cite{chen2014recognizing, li2014exploiting,sarafianos2017adaptive}. 
Vapnik \etal \cite{vapnik2009new} introduced the concept of privileged information, which is additional information that is only accessible during training. 
Subsequent works \cite{hoffman2016learning, lopez2015unifying, mordan2018revisiting, sharmanska2013learning} apply this principle to various adaptation tasks.
Entropy minimization has proven effective in semi-supervised learning \cite{grandvalet2004semi, springenberg2015unsupervised}. 
Vu \etal \cite{vu2019advent} were the first to introduce the technique for domain adaptation in semantic segmentation, using adversarial learning to optimize the entropy objective. 
Pan \etal \cite{pan2020unsupervised} built on this idea and proposed an intra-domain adaptation framework guided by the entropy levels of predictions. 
Their strategy consists of a two-stage process: first, adapting from the source to the target domain, and second, aligning the prediction distributions within the target domain by distinguishing between hard and easy samples. 
Another UDA approach is self-training, where a model's predictions are reused as pseudo-labels for training on unlabeled data. 
This strategy has been used in classification \cite{li2019learning} and segmentation tasks \cite{zou2018unsupervised}.
Meanwhile, bijective maximum likelihood was applied for domain adaptation \cite{truong2021bimal,truong2024conda}.
Truong \etal \cite{truong2023fredom} addressed the fairness in domain adaptation where the labels are imbalanced.
While the prior UDA methods show the advantages in general deep learning tasks, the lacks of physical properties hinder the models from achieving robustness in domain adaptation for 2D material identification tasks.

\section{Preliminary Background}

In this section, we first review the foundational principles of multilayer optical calculations, which play a crucial role in 2D material research by modeling the appearance of quantum flakes under microscopy. 
Next, we examine the imaging mechanism of microscopy and how it captures visual representations of quantum flakes. 
Based on this understanding, we demonstrate how synthetic quantum flake images (source domain) can be generated using physics-informed parameters derived from experimental setups. 
Finally, we discuss the construction of real quantum flake images (target domain) and highlight the key differences between these real images and their synthetic counterparts.

\subsection{Multilayer Optical Calculations}

\begin{wrapfigure}{r}{0.55\linewidth}
\vspace{-4mm}
\centering
\includegraphics[width=0.95\linewidth]{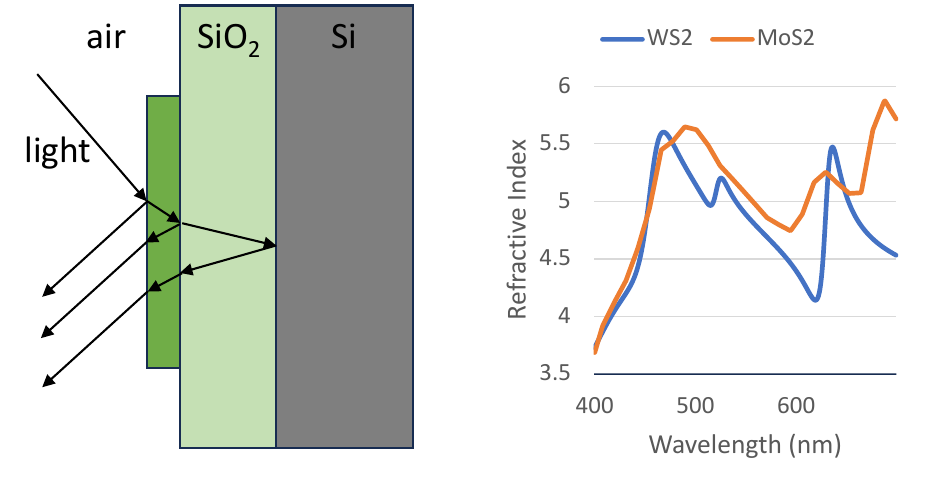}
\vspace{-4mm}
\caption{The optical model of a 2D material system.}
\label{fig:refractive_index}
\vspace{-2mm}
\end{wrapfigure}
We briefly review the optical calculations of multilayer films using the transfer matrix method to model light propagation through 2D material flakes. 
When light transitions between two stacked thin-layer materials $m_l$ and $m_{l+1}$, changes in refractive indices cause partial reflection $r_{l,l+1}$ and transmission $t_{l,l+1}$, described by the Fresnel equations.
Consider a multilayer thin film structure composed of $L \in \mathbb{N}$ distinct layers, indexed by $l \leq L$, the light enters the system through an injection layer of semi-infinite thickness, i.e., air, labeled $l = 0$, with a relative amplitude of $1$.
It exists through an outcoupling layer, also of semi-infinite thickness, labeled $l = L + 1$, from which no light re-enters the film.
The reflection and transmission of the entire multilayer film is formulated as in Eqn. \eqref{eq:reflection_trans}.
\begin{equation}
    \label{eq:reflection_trans}
    \begin{pmatrix}
        1 \\ v
    \end{pmatrix}
    = M_{0,1} \prod_{l=1}^L P_l M_{l,l+1}
    \begin{pmatrix}
        u \\ 0
    \end{pmatrix}
\end{equation}
where $v$ is the reflective intensity, $u$ is the forwarding intensity into the last layer $L+1$, $P_l$ is the wave evolution matrix through a layer $l$ and $M_{l,l+1}$ is the wave changing matrix between layer $l$ and $l+1$.
The detail of light propagation via the transfer matrix method is described in the Supplementary.
Figure \ref{fig:refractive_index} illustrates the optical model of a 2D material system. 
Typically, a 2D material system includes two layers of substrate made of Si and SiO$_2$, and a layer of flakes on top of the substrate. 
The flakes could be different types of material with different optical properties, as shown in Figure \ref{fig:refractive_index}.

\subsection{2D Material Image}

The 2D material image is captured using an optical microscope under white light consisting of a continuous spectrum of wavelengths from approximately 400\,nm to 700\,nm. 
These wavelengths pass through the air, interact with the 2D material and its substrate, and are partially reflected. The reflected light is collected by the microscope’s optical system and projected onto the sensor. The sensor measures the spectral response, depending on the incident illumination and the material’s spectral reflectance. To form the RGB image $x \in \mathbb{R}^{h \times w \times 3}$, the captured intensity at each pixel is computed by integrating the product of the sensor’s spectral sensitivity $S(\lambda)$, the illumination spectrum $I(\lambda)$, and the object’s reflectance $R(\lambda)$ over the visible wavelength range as in Eqn.~\eqref{eq:microscopy_image}.
\begin{equation}
    \label{eq:microscopy_image}
    x = \int S(\lambda) I(\lambda) R(\lambda) d\lambda \approx \sum_\lambda S(\lambda) I(\lambda) R(\lambda)
\end{equation}
In the discrete form, we can estimate image $x$ based on the transform 
$x = S^\top (I \circ R)$,
where $S \in \mathbb{R}^{D\times3}$, $I \in \mathbb{R}^D$, $R \in \mathbb{R}^D$, $D$ is the number of sampled wavelengths, and $\circ$ is the element-wise matrix multiplication.
\newline
\textbf{Synthesize image (source domain)}. We synthesize source domain images $x_s$ using known sensor and illumination parameters. 
Specifically, we adopt the CIE 1931 color matching functions \cite{CIE1931} for $S$ and the D65 standard illuminant \cite{CIE_D65} for $I$. 
To compute reflectance $R$, we specify material type (e.g., h-BN, MoS\textsubscript{2}), substrate (e.g., Si/SiO\textsubscript{2}), and substrate thickness (e.g., 290\,nm), then randomly generate flake shapes and thicknesses. 
The transfer matrix method shown in Eqn.~\eqref{eq:reflection_trans} is applied to obtain $R$. 
In particular, $x_s$ is measured as in Eqn.~\eqref{eq:synthesize_image}.
\begin{equation}
\label{eq:synthesize_image}
\begin{split}
    x_s = S_s^\top (I_s \circ R_s) = A_s R_s, \quad
    A_s = S_s^\top \text{diag} (I_s)
\end{split}
\end{equation}
% \newline
\textbf{Real image (target domain).} In theory, the real image can be formulated similarly to the synthesized image in Eqn.~\eqref{eq:synthesize_image}. However, it includes an additional factor $G$, representing the white balancing process. In practice, users often adjust these parameters on the microscope to improve the visibility or contrast of 2D material flakes. The white balance $G$ is sometimes personalized based on individual experimental setups in 2D quantum material discovery.
\begin{equation}
\begin{split}
    \label{eq:real_image}
    x_t = (S_t^\top (I_t \circ R_t)) \circ G_t = \text{diag}(G_t) A_t R_t, \quad
    A_t = S_t^\top \text{diag}(I_t)
\end{split}
\end{equation}

\textbf{Motivations}. To address the domain gap between $x_s$ and $x_t$, previous domain adaptation methods \cite{wei2022entropy,vs2023instance,hao2024simplifying,qiu2021source,sun2023domain,hwang2024sf} are generally unsuitable for 2D quantum material problems due to their purely statistical nature and lack of integration with physical principles.  
These methods aim to align feature distributions between source and target domains, assuming the gap is purely due to visual or statistical differences. 
However, in 2D quantum material identification, domain shift arises from underlying physical phenomena, e.g., variations in material thickness, interference patterns, refractive index, substrate types, and different microscopy, which are not easily captured by generic feature alignment strategies.
As a result, models trained with standard domain adaptation techniques often fail to generalize when applied to real images. 
This gap highlights the need for physics-informed adaptation, where material properties and optical modeling are explicitly used to guide the learning process, allowing better generalization and interpretability across real and synthetic domains.

\section{The Proposed Physics-Informed Adaptation ($\varphi$-Adapt) Approach}

\begin{figure}[t]
\centering
\includegraphics[width=\linewidth]{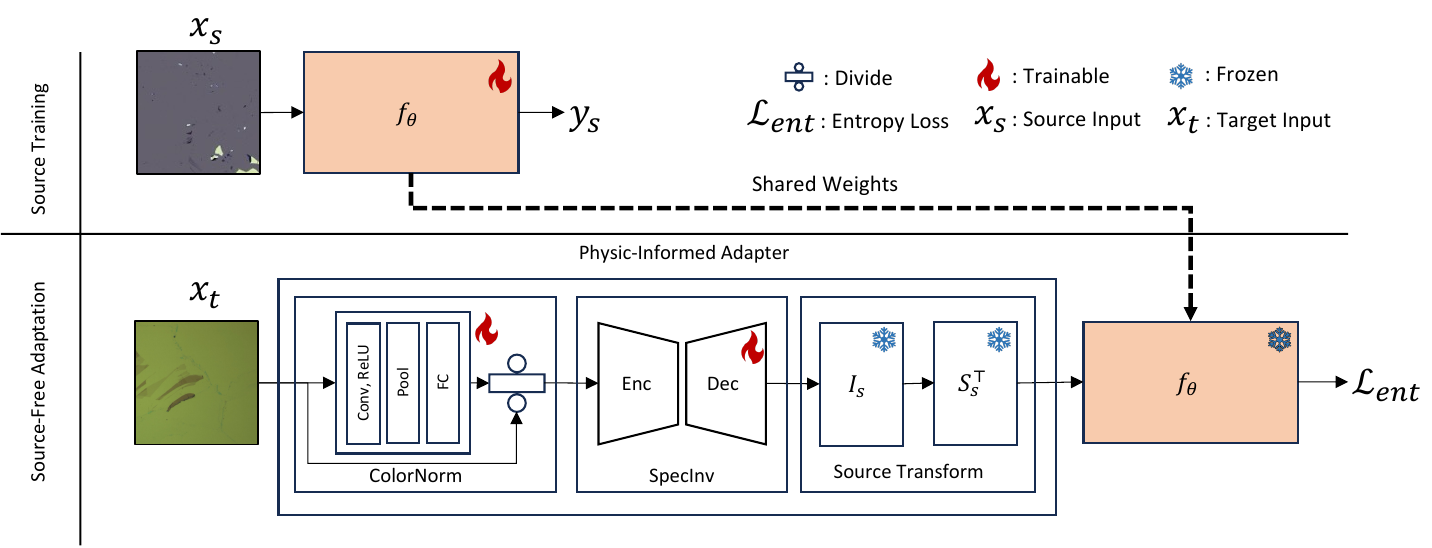}
\vspace{-4mm}
\caption{
The proposed Physics-Informed Adaptation framework. 
}
\vspace{-4mm}
\label{fig:overall}
\end{figure}

Our proposed approach is demonstrated in the Fig. \ref{fig:overall}. Firstly, we train the model w.r.t a specific task, such as detection, thickness estimation, etc, on the source data only. At the second step, we propose a physics-informed adaptation network to transfer the real image into the source style. The network contains three submodules, namely, ColorNorm, SpecInv, and the Source Transform module. The details of each module will be presented in the following sections.

\subsection{Training From Source Domain}
Let $\mathcal{D}_s = \{(x^s_i, y^s_i)\}_{i=1}^{N_s}$ be the set of samples in the source domain.
The $x \in \mathbb{R}^{h \times w \times c}$ is the RGB image, and $y$ is the task label, e.g., for thickness estimation or flake detection. We learn a model, denoted as $f_\theta$, on a given source dataset by minimizing the expected risk of the source-labeled data as in Eqn. \eqref{eq:expected_risk_source}.
\begin{align}
    \label{eq:expected_risk_source}
    \theta^{*} = \arg\min \mathbb{E}_{(x, y) \sim p_s(x, y)} \mathcal{L}_s(f_\theta(x, \theta), y) 
\end{align}
where $p_s(x, y)$ is a joint distribution of source data. $\mathcal{L}_s$ is the loss function while training in the source domain. In the flake detection task, $\mathcal{L}_s$ is simply cross-entropy, while in thickness estimation, we design $\mathcal{L}_s$ as the smooth-$L_1$.

\subsection{Physics-Informed Adaptation Network}
Assuming we can synthesize images $x_s$ using the same material configurations as in real-world experiments, the transformation matrix $A_s$ in Eqn.~\eqref{eq:synthesize_image} is fully known. Consequently, the domain gap between $x_s$ and $x_t$ arises from the unknown factors $G_t$ and $A_t$, where $G_t$ captures user-specific white balance adjustments, and $A_t$ reflects the hardware-dependent sensor and illumination characteristics. To close the gap between $x_s$ and $x_t$, it is necessary to determine these unknown factors. Once we know these factors, we can transform the $x_t$ to $x_s$ using Eqn. \eqref{eq:target_to_source}.
\begin{align}
    \label{eq:target_to_source}
    x_{t\rightarrow s} = A_s {R_t} = A_s (A_t^{-1}(G_t^{-1}(x_t)))
\end{align}
where ${R}_t  = A_t^{-1}(G_t^{-1}(x_t))$ is the estimated reflection factor of $x_t$. 
While $A_t^{-1}$ and $ G_t^{-1}$ are intractible, we propose novel learnable modules: \texttt{ColorNorm} $G_t^{-1}$ to estimate the color factors, e.g., white balance and gamma factors, and \texttt{SpecInv} $A_t^{-1}$ to estimate spectrum parameters.
 
\textbf{ColorNorm $G_t^{-1}$}. This module aims to predict the color factors $G_t \in \mathbb{R}^3$ w.r.t. input $x_t$, either the default settings of real-world microscopy or personalized by users. Even though $G_t$ is the global variable, the color distribution in an image varies locally. 
Thus, \texttt{ColorNorm} should be able to capture local chromatic contexts. 
Motivated by this idea, we design the \texttt{ColorNorm} module as a stack of \texttt{Conv}, \texttt{ReLU}, and \texttt{AvgPool} followed by fully connected \texttt{FC} layers. 
In summary, the color factors $G_t$ are measured as in Eqn. \eqref{eq:color_norm}.
\begin{align}
    \label{eq:color_norm}
    G_t &= \texttt{FC}(\texttt{AvgPool}(\texttt{ReLU}(\texttt{Conv}(x_t)))
\end{align}
The color normalization image, or raw image, of $x_t$, denoted as $\bar{x_t}$, is calculated as: $\bar{x_{t}} = x_t / G_t$. 

\noindent
\textbf{SpecInv \( A_t^{-1} \).}  
From Eqn.~\eqref{eq:real_image}, we observe that the real image can be represented as
$\bar{x}_t = A_t R_t$, where $R_t$ denotes the reflection factors of $x_t$. These reflection factors are determined by the material configuration rather than the hardware setup. Assuming we can synthesize an image $x_s$ with the same material properties as in the real-world scenario, it implies that a source-style image can be generated from $R_t$. However, estimating $R_t$ is a challenging task due to the intractability of $A_t$. To overcome this, we introduce a learnable module named \texttt{SpecInv}, which aims to recover $R_t$ from $\bar{x}_t$. Notably, $R_t$ is a spatial map where each element corresponds to the reflection intensity at a pixel location, reflecting the material stack's influence on light. Thus,
$R_t \in \mathbb{R}^{H \times W \times 1}$ shares the same spatial resolution as $\bar{x}_t \in \mathbb{R}^{H \times W \times 3}$, but contains only a single channel. Based on this structure, we design the \texttt{SpecInv} module as an encoder-decoder architecture tailored to predict $R_t$ from the observed image $\bar{x}_t$ as follows,
\begin{align}
    R_t = \texttt{SpecInv} (x_t) = \texttt{Decoder}(\texttt{Encoder}(x_t)).
\end{align}

\textbf{Source Transform}. Given the estimated $R_t$ from the \texttt{SpecInv} module, the source-style image $x_{t \rightarrow s}$ can be generated according to Eqn.~\eqref{eq:target_to_source}. Specifically, we compute $x_{t \rightarrow s}$ by multiplying $R_t$ with the known illumination spectrum $I_s$, followed by the transpose of the spectral sensitivity matrix $S_s^\top$.

\subsection{Physics-Based Learning with Synthesized Dataset}

\begin{figure}
\centering
\includegraphics[width=0.95\linewidth]{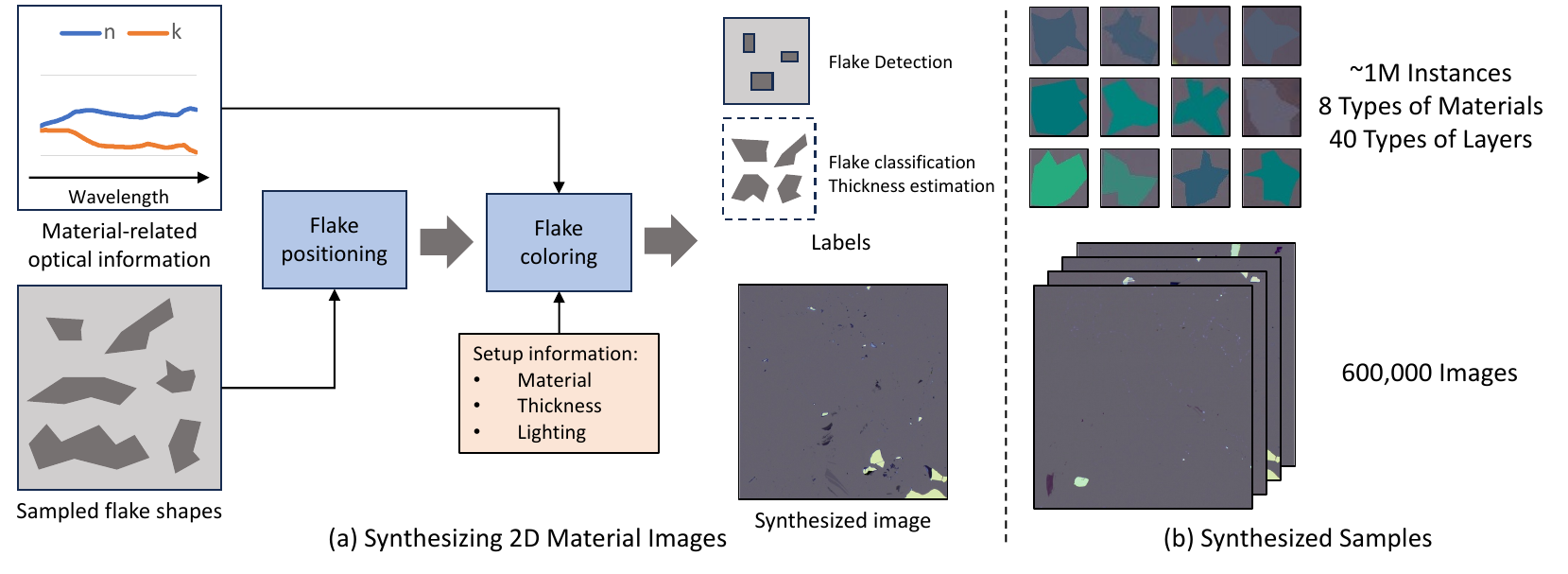}
\vspace{-4mm}
\caption{
(a) The process of generating the synthesized 2D material dataset. 
(b) The synthesized dataset includes 600,000 images with approximately one million instances.
}
\vspace{-4mm}
\label{fig:flake_synthesized_data.pdf}
\end{figure}

To precisely adapt the target images, the parameters of the proposed physics-based material shifting module should be well-represented.
As the physics-based material shifting module contains reflective information of the material and lighting, precise physical parameters and data labeling are required.
The 2D materials collecting and labeling process is time-consuming, costly, and not precise.
Hence, we introduce a synthesized dataset for various 2D material analysis tasks, including labeling and optical information.
The synthesized dataset is generated from reflective information of different materials and lighting conditions. 
Thus, we can precisely control the physical information of the source domain.
The synthesized dataset can be utilized for two training steps, including optical representation learning and specific-task training.

\textbf{Optical representation learning.}
As the optical parameters, i.e., material types, refractive indices, material thicknesses, and lighting conditions, are included in the synthesized dataset, they can be directly applied to the physics-based material shifting module.

\textbf{Task-specific training.}
With the large scale of synthesized data, the 2D material model $f_\theta$ can be well-trained for multiple tasks, i.e., flake detection, flake classification, and thickness estimation.
Let $\hat{x}$ be the reconstructed image computed from the physics-based material shifting module, the 2D material model is trained to predict a task-specific output $\hat{y}$ via task-specific loss function $\mathcal{L}_\text{task}(\hat{y}, \bar{y})$, where $\bar{y}$ is the ground-truth.

\subsection{Source-Free Entropy Minimization}

Entropy minimization is a widely used strategy in domain adaptation to encourage confident predictions on unlabeled target domain data.
The core intuition is that a well-adapted model should produce low-entropy, i.e., high-confidence, output when applied to target samples.
Given a model output $p(\hat{y}|x_t)$ for a target sample $x_t$, the entropy of the prediction is defined as $H(\hat{y}) = -\sum_{c=1}^C p(\hat{y} = c | x_t) \log p(\hat{y} = c | x_t)$, where $C$ is the number of classes.
By minimizing the average entropy across the target data, the model is encouraged to align the target features with source class clusters.

To adapt the target 2D material images set, we minimize the entropy to update the Physics-Informed Adaptation Network.
First, the optimizer collects the parameters $\theta_\texttt{ColorNorm}$ of \texttt{ColorNorm} module and $\theta_\texttt{SpecInv}$ of \texttt{SpecInv} module, which are initially obtained from the source data.
The parameters of \texttt{ColorNorm} and \texttt{SpecInv} are updated via the gradient of the prediction entropy $\nabla H(\hat{y})$.

As $\theta_\texttt{SpecInv}$ represents the optical properties of $x$, including the refractive index of each wavelength $\lambda$, the parameters $\theta_\texttt{SpecInv}$ should be well-structured.
For two neighboring wavelengths $\lambda$ and $\lambda^\prime$, the optical information should be nearly similar.
Thus, we present a pair-wise regularization $\tau_\text{neighbor}(\theta_\texttt{SpecInv})$ to encourage the optical similarity between $\lambda$ and $\lambda^\prime$ as follows:
\begin{equation}
    \tau_\text{neighbor}(\theta_\texttt{SpecInv}) = 
    \frac{1}{D} \sum_\lambda \sum_{\lambda^\prime \in \mathcal{N}_\lambda} \left(
    || \theta_\texttt{SpecInv}^{(\lambda)} - \theta_\texttt{SpecInv}^{(\lambda^\prime)} ||_2^2
    \right)
\end{equation}
where $D$ is the number of sampled wavelengths, $\mathcal{N}_\lambda$ is the set of neighboring wavelengths of $\lambda$, and $\theta_\texttt{SpecInv}^{(\lambda)}$ is the inverse optical parameters at wavelength $\lambda$.

\section{Experiments}

\subsection{Implementation Details}

The physics-based material shifting module samples $D = 128$ wavelengths uniformly, ranging from 380 nm to 780 nm.
Our approach is implemented in PyTorch and trained on eight NVIDIA RTX A6000 GPUs with 48 GB of VRAM. 
The model is optimized by the Adam optimizer \cite{kingma2014adam} with momentum $0.9$, weight decay $10^{-4}$, and batch sizes of 2 per GPU for the flake detection task and 8 per GPU for the flake classification and thickness estimation tasks.

\subsection{Datasets and Benchmarks}

\begin{table}[t]
\centering
\caption{Flake detection results on MoS$_2$ material of Masubuchi \etal~benchmark \cite{masubuchi2020deep}.}
\vspace{-2mm}
\resizebox{0.86\linewidth}{!}{
\begin{tabular}{lllccc}
\Xhline{2\arrayrulewidth}
\textbf{Method} & \textbf{Detector} & \textbf{Backbone} & \textbf{AP (\%)} & \textbf{AP$^{.50}$ (\%)} & \textbf{AP$^{.75}$ (\%)} \\
\hline
Source only & MaskRCNN \cite{he2017mask} & ResNet-50 \cite{he2016deep} & 16.7 & 32.6 & 17.1 \\
Source only & ViTDet \cite{li2022exploring} & ViT-Base \cite{dosovitskiy2020image} & 14.1 & 28.3 & 14.2 \\
Source only & SAM \cite{kirillov2023segment} & ViT-Base \cite{dosovitskiy2020image} & 17.7 & 33.9 & 18.2 \\
Source only & SAMv2 \cite{ravi2024sam} & ViT-Base \cite{dosovitskiy2020image} & 18.9 & 34.1 & 18.6 \\
\hline
Wei \etal \cite{wei2022entropy} & MaskRCNN \cite{he2017mask} & ResNet-50 \cite{he2016deep} & 28.1 & 47.6 & 29.5 \\
IRG \cite{vs2023instance} & MaskRCNN \cite{he2017mask} & ResNet-50 \cite{he2016deep} & 29.6 & 46.8 & 27.6 \\
Hao \etal \cite{hao2024simplifying} & MaskRCNN \cite{he2017mask} & VGG16 \cite{simonyan2014very} & 30.2 & 48.1 & 29.8 \\
Hao \etal \cite{hao2024simplifying} & MaskRCNN \cite{he2017mask} & ResNet-50 \cite{he2016deep} & 29.7 & 47.4 & 30.1 \\
\hline
Wei \etal \cite{wei2022entropy} & ViTDet \cite{li2022exploring} & ViT-Base \cite{dosovitskiy2020image} & 27.9 & 48.4 & 30.2 \\
IRG \cite{vs2023instance} & ViTDet \cite{li2022exploring} & ViT-Base \cite{dosovitskiy2020image} & 28.2 & 49.3 & 29.7 \\
Hao \etal \cite{hao2024simplifying} & ViTDet \cite{li2022exploring} & ViT-Base \cite{dosovitskiy2020image} & 26.0 & 46.6 & 28.1 \\
\hline
\textbf{$\varphi$-Adapt (Ours)} & \textbf{MaskRCNN} \cite{he2017mask} & \textbf{ResNet-50} \cite{he2016deep} & \textbf{34.1} & \textbf{51.2} & \textbf{32.2} \\
\textbf{$\varphi$-Adapt (Ours)} & \textbf{ViTDet} \cite{li2022exploring} & \textbf{ViT-Base} \cite{dosovitskiy2020image} & \textbf{32.5} & \textbf{50.4} & \textbf{31.3} \\
\Xhline{2\arrayrulewidth}
\end{tabular}
}
\vspace{-2mm}
\label{tab:flake_detection_masubuchi}
\end{table}     

\begin{table}[t]
\centering
\caption{Flake layer classification results on Masubuchi \etal~benchmark \cite{masubuchi2020deep}.}
\vspace{-2mm}
\resizebox{0.95\linewidth}{!}{
\begin{tabular}{llcccc}
\Xhline{2\arrayrulewidth}
\textbf{Method} & \textbf{Backbone} & \makecell{\textbf{h-BN} \\ \textbf{Accuracy (\%)}} & \makecell{\textbf{Graphene} \\ \textbf{Accuracy (\%)}} & \makecell{\textbf{MoS$_2$} \\ \textbf{Accuracy (\%)}} & \makecell{\textbf{WTe$_2$} \\ \textbf{Accuracy (\%)}} \\
\hline
Source only & ResNet-50 \cite{he2016deep} & 80.5 & 41.4 & 66.2  & 57.4 \\
Source only & ViT-Base \cite{dosovitskiy2020image} & 63.4 & 46.3 & 72.7 & 40.7 \\
\hline
CPGA \cite{qiu2021source} & ResNet-50 \cite{he2016deep} & 82.9 & 78.3 & 77.5 & 83.3 \\
Sun \etal \cite{sun2023domain} & ResNet-50 \cite{he2016deep} & 81.7 & 86.2 & 76.0 & 85.2 \\
SF(DA)$^2$ \cite{hwang2024sf} & ResNet-50 \cite{he2016deep} & 87.8 & 74.3 & 85.7 & 81.3 \\
\hline
CPGA \cite{qiu2021source} & ViT-Base \cite{dosovitskiy2020image} & 84.1 & 76.4 & 80.5 & 82.3 \\
Sun \etal \cite{sun2023domain} & ViT-Base \cite{dosovitskiy2020image} & 89.0 & 79.4 & 81.8 & 83.7 \\
SF(DA)$^2$ \cite{hwang2024sf} & ViT-Base \cite{dosovitskiy2020image} & 85.4 & 79.0 & 85.0 & 82.8 \\
\hline
\textbf{$\varphi$-Adapt (Ours)} & \textbf{ResNet-50} \cite{he2016deep} & \textbf{93.9} & \textbf{87.6} & \textbf{89.6} & \textbf{86.1} \\
\textbf{$\varphi$-Adapt (Ours)} & \textbf{ViT-Base} \cite{dosovitskiy2020image} & \textbf{92.7} & \textbf{86.7} & \textbf{90.9} & \textbf{86.6} \\
\Xhline{2\arrayrulewidth}
\end{tabular}
}
\vspace{-4mm}
\label{tab:flake_classification_masubuchi}
\end{table}

\noindent
\textbf{Datasets:}
\textbf{Masubichi \etal} \cite{masubuchi2020deep} is a flake detection dataset that consists of 6,833 training samples and 1,705 testing samples of four types of materials, i.e., h-BN, graphene, MoS$_2$, and WTe$_2$.
\textbf{Uslu \etal} \cite{uslu2024open} is a 2D material dataset including 517 training samples and 1,782 testing samples of two types of materials, i.e., graphene and WSe$_2$.
Moreover, we also collected 177 samples of flakes, with a range of thickness from 10nm to 240nm, for thickness estimation.

\noindent
\textbf{Benchmarks:} We focus on three 2D flake identification tasks, including flake detection, flake layer classification, and thickness estimation.
\textbf{Flake detection} localizes the specific flakes required in each dataset.
\textbf{Flake layer classification} distinguishes the type of the detected flakes, e.g., one layer, two layers, or three layers.
\textbf{Thickness estimation} predicts the thickness of the detected flakes.

\subsection{Comparisons with Prior SOTA Methods}

\begin{wraptable}[12]{r}{0.5\linewidth}
\vspace{-6mm}
\centering
\caption{Flake layer classification results on Uslu \etal~benchmark \cite{uslu2024open}.}
\resizebox{\linewidth}{!}{
\begin{tabular}{llcc}
\Xhline{2\arrayrulewidth}
\textbf{Method} & \textbf{Backbone} & \makecell{\textbf{Graphene} \\ \textbf{Acc (\%)}} & \makecell{\textbf{MoS$_2$} \\ \textbf{Acc (\%)}} \\
\hline
Source only & ResNet-50 \cite{he2016deep} & 31.2 & 40.3 \\
Source only & ViT-Base \cite{dosovitskiy2020image} & 34.9 & 48.8 \\
\hline
CPGA \cite{qiu2021source} & ResNet-50 \cite{he2016deep} & 38.6 & 77.3 \\
Sun \etal \cite{sun2023domain} & ResNet-50 \cite{he2016deep} & 35.2 & 74.8 \\
SF(DA)$^2$ \cite{hwang2024sf} & ResNet-50 \cite{he2016deep} & 44.6 & 79.3 \\
\hline
CPGA \cite{qiu2021source} & ViT-Base \cite{dosovitskiy2020image} & 44.4 & 79.6 \\
Sun \etal \cite{sun2023domain} & ViT-Base \cite{dosovitskiy2020image} & 45.9 & 78.5 \\
SF(DA)$^2$ \cite{hwang2024sf} & ViT-Base \cite{dosovitskiy2020image} & 46.6 & 80.5 \\
\hline
\textbf{$\varphi$-Adapt (Ours)} & \textbf{ResNet-50} \cite{he2016deep} & \textbf{47.0} & \textbf{86.2} \\
\textbf{$\varphi$-Adapt (Ours)} & \textbf{ViT-Base} \cite{dosovitskiy2020image} & \textbf{52.0} & \textbf{86.6} \\
\Xhline{2\arrayrulewidth}
\end{tabular}
}
\label{tab:flake_classification_uslu}
\end{wraptable}
% \noindent
\textbf{Masubuchi \etal~flake detection.}
The benchmark focuses on single-class flake detection.
Table \ref{tab:flake_detection_masubuchi} describes the experimental results on the MoS$_2$ of the proposed approach compared to the previous state-of-the-art methods.
In detail, the average precision (AP) of our approach is $34.1\%$ using the convolutional backbone, i.e., ResNet-50 \cite{he2016deep}.
Meanwhile, the proposed approach with the ViT-Base backbone achieves the AP of $32.5\%$, higher than prior methods in the same backbone.

\noindent
\textbf{Masubuchi \etal~flake classification.} 
The target flakes have a number of layers from 1 to 10. They are grouped into three classes, including monolayer, having one layer, fewlayer, having the number of layers ranging from two to nine layers, and thicklayer, having ten layers.
As shown in Table \ref{tab:flake_classification_masubuchi}, the proposed approach outperforms previous methods \cite{qiu2021source,sun2023domain,hwang2024sf} in multiple backbones \cite{he2016deep,dosovitskiy2020image} by a large margin.
Compared on ResNet-50 \cite{he2016deep}, our approach shows better performance with the accuracies of $93.9\%$, $87.6\%$, $89.6\%$, and $86.1\%$ on the h-BN, graphene, MoS$_2$, and WTe$_2$ materials, respectively.
Meanwhile, the accuracies of the proposed approach using the vision transformer model \cite{dosovitskiy2020image} are $92.7\%$, $86.2\%$, $90.9\%$, and $86.6\%$.

\noindent
\textbf{Uslu \etal~flake classification.}
The benchmark has different numbers of target layers in the graphene and MoS$_2$ materials.
The graphene material has several target layers ranging from one to four layers.
Meanwhile, the MoS$_2$ material has several target layers in the range of one to three layers.
Table \ref{tab:flake_classification_uslu} illustrates the experimental results of our approach compared to prior methods.
In detail, the proposed approach achieves the accuracies of $47.0\%$ and $52.0\%$ using ResNet-50 \cite{he2016deep} and ViT-Base \cite{dosovitskiy2020image} on the graphene material.
Meanwhile, the accuracies on the MoS$_2$ are $86.2\%$ and $86.6\%$.

\begin{wraptable}[8]{r}{0.45\linewidth}
\vspace{-6mm}
\centering
\caption{Thickness estimation comparisons on our collected flakes.}
\resizebox{\linewidth}{!}{
\begin{tabular}{llc}
\Xhline{2\arrayrulewidth}
\textbf{Method} & \textbf{Backbone} & \textbf{Error (nm)} \\
\hline
Source only & ResNet-50 \cite{he2016deep} &  25.3 \\
\hline
CPGA \cite{qiu2021source} & ResNet-50 \cite{he2016deep} & 18.2 \\
Sun et al. \cite{sun2023domain} & ResNet-50 \cite{he2016deep} & 17.8 \\
SF(DA)$^2$ \cite{hwang2024sf} & ResNet-50 \cite{he2016deep} & 14.9 \\
\hline
\textbf{$\varphi$-Adapt (Ours)} & \textbf{ResNet-50} \cite{he2016deep} & \textbf{5.8} \\
\Xhline{2\arrayrulewidth}
\end{tabular}
}
\label{tab:thickness_estimation}
\end{wraptable}
\noindent
\textbf{Thickness Estimation.}
To demonstrate the robustness of our proposed approach, we collect and measure real flake instances to evaluate thickness estimation.
In general, the domain adaptation for thickness estimation has two folds.
First, we quantize the thicknesses of the flakes and adapt the model similarly to flake classification.
Then, we train a new linear regression head upon the trained backbone for the thickness estimation.
Table \ref{tab:thickness_estimation} illustrates the experimental results of our approach compared to prior methods \cite{qiu2021source,sun2023domain,hwang2024sf}.
In detail, the proposed approach achieves an error of $5.8$nm, $9.1$nm less than the previous domain adaptation methods.

\subsection{Ablation Studies}

\begin{wraptable}[10]{r}{0.45\linewidth}
\vspace{-8mm}
\centering
\caption{Effectiveness of our approach on Graphene material of Masubuchi \etal~flake classification benchmark \cite{masubuchi2020deep}.}
\resizebox{0.95\linewidth}{!}{
\begin{tabular}{cccccc}
\Xhline{2\arrayrulewidth}
\makecell{\textbf{Color} \\ \textbf{Norm}} & \makecell{\textbf{Source} \\ \textbf{Transform}} & $\mathcal{L}_\text{ent}$ & $\tau_\text{neightbor}$ & \makecell{\textbf{Accuracy} \\ (\%)} \\
\hline
  &  &  &  & 41.4 \\
  &  & \checkmark &  & 73.5 \\
\hline
  & \checkmark & \checkmark &  & 80.2 \\
  & \checkmark & \checkmark & \checkmark & 83.4 \\
\hline
 \checkmark &  & \checkmark &  & 79.7 \\
 \checkmark & \checkmark & \checkmark &  & 84.3 \\
 \checkmark & \checkmark & \checkmark & \checkmark & 87.6 \\
\Xhline{2\arrayrulewidth}
\end{tabular}
}
\label{tab:ablation}
\end{wraptable}
Our ablation experiments study the effectiveness of our proposed approach on the Graphene material of Masubuchi \etal~\cite{masubuchi2020deep} classification benchmark as shown in Table \ref{tab:ablation}.

\noindent
\textbf{Effectiveness of physics-informed adaptation modules.}
We evaluate the impact of physics-informed adaptation modules, including color normalization and source transformation.
As shown in Table \ref{tab:ablation}, color normalization predicts the white balance of the target distribution, improving the model performance.
In detail, the accuracy is improved from $73.5\%$ to $79.7\%$.
Meanwhile, source transformation shifts the target optical distribution to the source optical distribution.
The accuracies without and with color normalization are increased from $73.5\%$ and $79.7\%$ to $80.2\%$ and $84.3\%$, respectively.

\noindent
\textbf{Effectiveness of losses.}
As reported in Table \ref{tab:ablation}, entropy loss helps the model to learn the target distribution and improves the accuracy from $41.4\%$ to $73.5\%$.
Meanwhile, neighbor regularization structures the target refractive index to make the model robust to the target optical distribution.
In detail, the accuracies without and with color normalization are increased from $80.2\%$ and $84.3\%$ to $83.4\%$ and $87.6\%$, respectively.

\section{Conclusions}

This paper has introduced $\varphi$-Adapt, a new physics-informed domain adaptation approach for 2D material identification.
Motivated by the optical properties of 2D materials, we have proposed a novel Physics-Informed Adaptation Network to learn and adapt the target distribution effectively.
Then, to adapt the distribution of the target images precisely, we have introduced a new synthesized 2D materials dataset including optical information for physics-based learning.
Finally, we have presented source-free entropy minimization to optimize the Physics-Informed Adaptation modules.
Our experimental results have demonstrated the improvement of the proposed physics-informed domain adaptation approach.
Our work bridges the gap between physics-based modeling and domain adaptation, enabling the impact of using synthesized data in real-world 2D material identification. 

\noindent
\textbf{Limitations:} While the proposed approach shows the robustness in physics-informed adaptation from the synthesized dataset to real datasets, other aspects besides optical properties are not considered, including sensor noises and artifacts.
Future work could explore these factors to enhance the performance in the noisy real-world data.

\noindent
\textbf{Broader Impacts:} Our contributions enhance the reliability and accuracy of automated quantum flake identification while having low real data.
By promoting a more physically grounded approach to domain adaptation, our approach has the potential to advance the field of 2D material analysis and the broader application of deep learning in scientific research.

\bibliographystyle{abbrv}
\bibliography{references}

\medskip

%%%%%%%%%%%%%%%%%%%%%%%%%%%%%%%%%%%%%%%%%%%%%%%%%%%%%%%%%%%%

% \input{supplementary}

% \input{checklist}

%%%%%%%%%%%%%%%%%%%%%%%%%%%%%%%%%%%%%%%%%%%%%%%%%%%%%%%%%%%%

\end{document}